\documentclass[runningheads]{llncs}
\usepackage{graphicx}
\usepackage{amsmath,amssymb} %
\usepackage{color}
\usepackage[width=122mm,left=12mm,paperwidth=146mm,height=193mm,top=12mm,paperheight=217mm]{geometry}
\usepackage{bm}
\usepackage{sidecap}
\DeclareMathOperator*{\argmin}{\arg\!\min}
\DeclareMathOperator*{\argmax}{\arg\!\max}

\newcommand{\eg}{\emph{e.g.}}
\newcommand{\ie}{\emph{i.e.}}
\newcommand{\etal}{\emph{et al.}}
\newcommand{\vs}{\emph{vs.}}
\def\w{{\mathbf w}}
\def\feature{{\boldsymbol \phi}}

\begin{document}
\pagestyle{headings}
\mainmatter

\title{Image Co-localization by Mimicking a Good Detector's Confidence Score Distribution\thanks{First two authors contributed equally. Correspondence
  should be addressed to C. Shen (e-mail: {\tt chunhua.shen@adelaide.edu.au}).
}
}

\titlerunning{Image Co-localization by Mimicking a Good Detector's Distribution}

\authorrunning{Y. Li, L. Liu, C. Shen and A. van den Hengel}

\author{Yao Li, Lingqiao Liu, Chunhua Shen,
  Anton van den Hengel}

\institute{School of Computer Science, The University of Adelaide, Australia
}

\maketitle

\begin{abstract}
Given a set of images containing objects from the same category, the task of image co-localization is to identify and localize each instance. This paper shows that this problem can be solved by
a simple but intriguing idea, that is, a common object detector can be
learnt by making its detection confidence scores distributed
like those of a strongly supervised detector. More specifically, we
observe that given a set of object proposals extracted from an image that
contains the object of interest, an accurate strongly supervised object
detector should give high scores to only a small minority of proposals,
and low scores to most of them. Thus, we devise an entropy-based objective function to enforce the above property when learning the common object detector. Once the detector is learnt, we resort to a segmentation approach to refine the localization. We show that despite its simplicity, our approach outperforms state-of-the-art methods.
\end{abstract}

\section{Introduction}
\label{sec:intro}

There has been an explosion of images available on the Internet in recent years, largely due to the popularity of photo sharing sites like Facebook and Flicker.
However, most of these images are either unlabeled or weakly-labeled.
One way of accessing these images is finding images depicting the same object, for instance, Google Image Search will return the images contains the common object described by the user input keyword.
In this paper, we aim to localize the common object in the scenario like this, \ie,  without using any other forms of supervision (manually-labeled bounding boxes or negative images). This task is known as the image co-localization task~\cite{DBLP:conf/cvpr/TangJLF14,DBLP:conf/eccv/JoulinTF14,DBLP:conf/cvpr/ChoKSP15} in literature.

Image co-localization is a particularly challenging task, and thus there exist a limited number of comparable methods~\cite{DBLP:conf/cvpr/TangJLF14,DBLP:conf/eccv/JoulinTF14,DBLP:conf/cvpr/ChoKSP15}.
These methods address this problem from various perspectives. The work in~\cite{DBLP:conf/cvpr/TangJLF14} introduces binary latent variables to indicate the presence of the common object and formulates the co-localization via latent variable inference.
The work of~\cite{DBLP:conf/cvpr/ChoKSP15}, in contrast, localizes the common object by matching common object parts.
Our work differs from previous approaches in that it directly learns the common object detector by modeling its detection confidence score distribution on each image, and achieves the localization with the learned detector.

The key insight in our method is that although we do not have sufficient supervision to learn a strongly supervised object detector, it is still possible to learn a detector by enforcing its detection confidence scores distributed as those of a strongly supervised detector. For a strongly supervised object detector, we have made the following observation: when an accurate strongly supervised object detector is applied to an image contains the object of interest, only a small minority of proposals are given high detection confidence scores while most of them are associated with low scores. Motivated by the key insight and the above observation, in this paper we design a novel Shannon-entropy-based objective function to promote the scarcity of high detection confidence scores within an image while avoiding the trivial solution of producing low scores for all proposals. In other words, by optimizing the proposed objective, our approach will encourage the existence of a few high response proposals in each image as the common object while suppressing responses in the remainder proposals which will be deemed as background.

To generate the final co-localization results, we have also devised a method for improving the bounding box estimate.
Inspired by detection-by-segmentation approaches (\eg, \cite{DBLP:conf/iccv/ParkhiVJZ11}), we use the final detection heat map
and color information to define a CRF-based segmentation algorithm, the output of which indicates the
instances of the common object.

Through an extensive evaluation on several benchmark datasets,
including the PASCAL VOC 2007 and 2012~\cite{DBLP:journals/ijcv/EveringhamEGWWZ15}, and also some subsets of
the ImageNet~\cite{DBLP:conf/cvpr/DengDSLL009},
we demonstrate that our approach not only outperforms the state-of-the-art in image co-localization, but is also on par with some weakly supervised
object localization approaches.

\section{Related Work}
\label{sec:related_work}
Image co-localization shares some similarities with image co-segmentation~\cite{DBLP:conf/cvpr/JoulinBP10,DBLP:conf/cvpr/RubinsteinJKL13,DBLP:conf/cvpr/ChenSG14} in the sense that both problems require a set of images of objects from a common category as input.
Instead of generating a precise segmentation of the related objects in each image, co-localization algorithms~\cite{DBLP:conf/cvpr/TangJLF14,DBLP:conf/eccv/JoulinTF14,DBLP:conf/cvpr/ChoKSP15} aim to draw a tight bounding box around the object.
Image co-localization is also related to works on weakly supervised object localization (WSOL)~\cite{DBLP:conf/iccv/SivaX11,DBLP:journals/ijcv/DeselaersAF12,DBLP:conf/cvpr/CinbisVS14,DBLP:conf/iccv/ShiHX13,DBLP:conf/eccv/WangRHT14,DBLP:conf/cvpr/BilenPT15,DBLP:conf/iccv/WangZYB15,DBLP:journals/pami/RenHTT16} as both try to localize objects of the same type within an image set, the key difference is WSOL requires manually-labeled negative images whereas co-localization does not. 

Tang \etal~\cite{DBLP:conf/cvpr/TangJLF14} formulate co-localization as a boolean constrained quadratic program which can be relaxed to a convex problem, which is further accelerated by the Frank-Wolfe Algorithm~\cite{DBLP:conf/eccv/JoulinTF14}.
Recently, Cho \etal~\cite{DBLP:conf/cvpr/ChoKSP15} propose a Probabilistic Hough Matching algorithm to match object proposals across images and then dominant objects are localized by selecting proposals based on matching scores.
There are also approaches address the problem of co-localization in video~\cite{DBLP:conf/cvpr/PrestLCSF12,DBLP:conf/eccv/JoulinTF14,DBLP:conf/iccv/KwakCPSL15}.
Notably, Prest~\etal~\cite{DBLP:conf/cvpr/PrestLCSF12} select spatio-temporal tubes which are likely to contain the common object, and
Joulin~\etal~\cite{DBLP:conf/eccv/JoulinTF14}, in contrast, extend~\cite{DBLP:conf/cvpr/TangJLF14} by incorporating temporal consistency.

However, in this paper, we tackle the co-localization problem from a new perspective, that is, learning the common object detector by modeling its detection confidence score distribution, and thus get rid of the need of 
manually-labeled negative images. 
An advantage of the proposed approach for learning common object detectors is that it provides an explicit mechanism by which to exploit the relationship between localization and detection.
The benefits of exploiting this relationship have been identified before in WSOL.
In~\cite{DBLP:journals/ijcv/DeselaersAF12}, objects are localized by minimizing a Conditional Random Field (CRF) energy function which incorporates class-specific information, and the class-specific information is learned from the localized objects.
Cinbis~\etal~\cite{DBLP:conf/cvpr/CinbisVS14} propose a multi-fold training procedure for Multiple Instance Learning
whereby, at each iteration, positive instances in each fold are localized by a detector trained from other folds in the previous iteration.
The approach that we propose here,
however, is the first to systematically leverage the idea of jointly performing object detection and localization for co-localizing common objects in images.

\section{Approach}
\label{sec:approach}

\begin{figure*}[t]
\begin{center}
\includegraphics[width=1\linewidth]{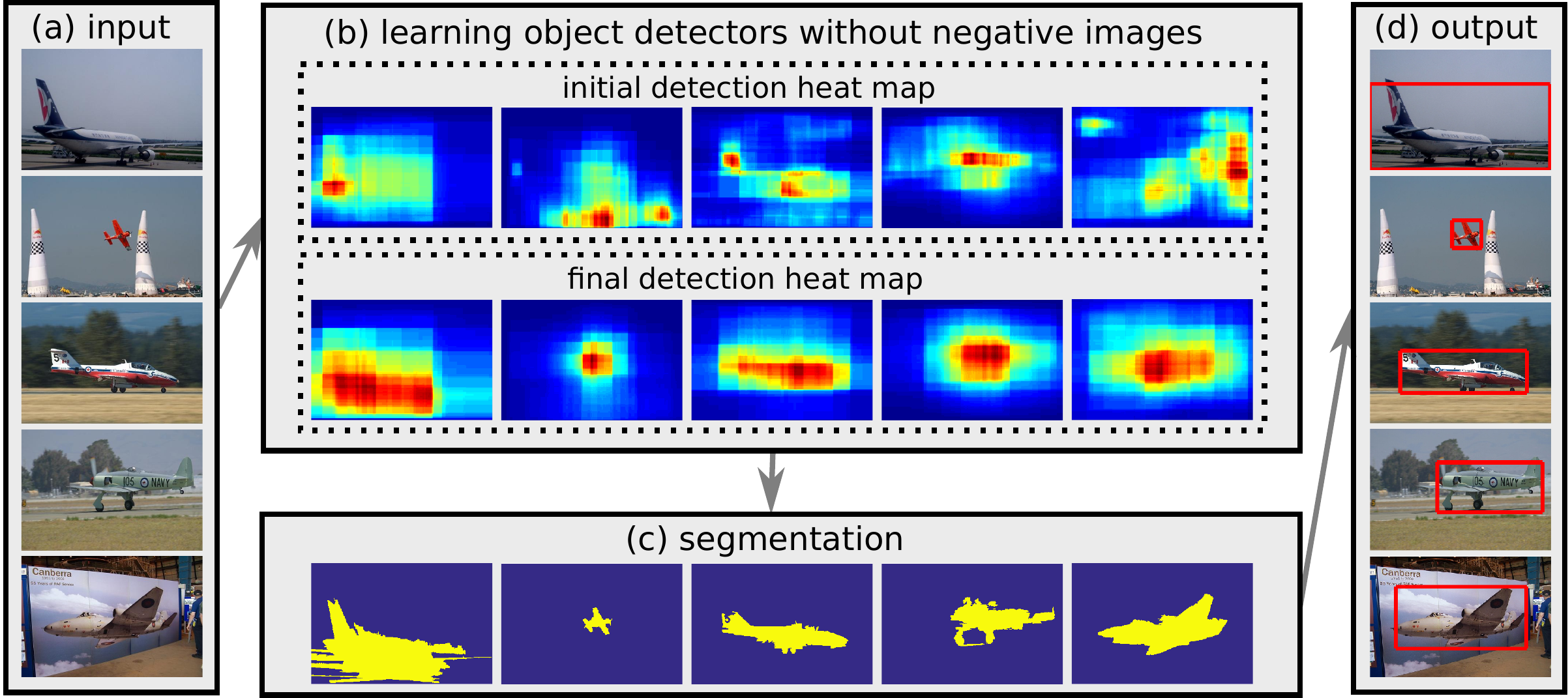} \\
\end{center}
\caption{An overview of our image co-localization framework. (a) The input of our system  is a set of images contains a common object category (here, aeroplane). (b) The common object detector is learnt by modeling the distribution of detection confidence scores.
(c) Detection heat maps generated by the learnt detector are used as the unary potential for graph-cuts segmentation.
(d) The output for each image is the smallest rectangle which covers the  corresponding segmentation.
}
\label{fig:overview} %
\end{figure*}

We give an overview of our image co-localization framework in Fig.~\ref{fig:overview}.
The input to our framework is a set of $N$ images $\mathcal{I}=\{I_1,I_2,\ldots,I_N\}$ contains one common object (\eg, aeroplane), and we aim to annotate the location of common object instances in each image.
Inspired by the behaviour of an accurate strongly supervised object detector (Sec.~\ref{subsec:long-tail}), the core of our framework is the procedure of learning the common object detector by modeling its detection confidence score distribution (Sec.~\ref{subsec:heatmap}).  
We further formulate object localization as a segmentation problem (Sec.~\ref{subsec:segmentation}), which involves using the detection heat map to define unary potentials of a binary energy function and solve it efficiently by standard graph-cuts.

\begin{SCfigure*}[][h]
\centering
\begin{tabular}{@{}c@{}c}
\includegraphics[width=0.32\textwidth]{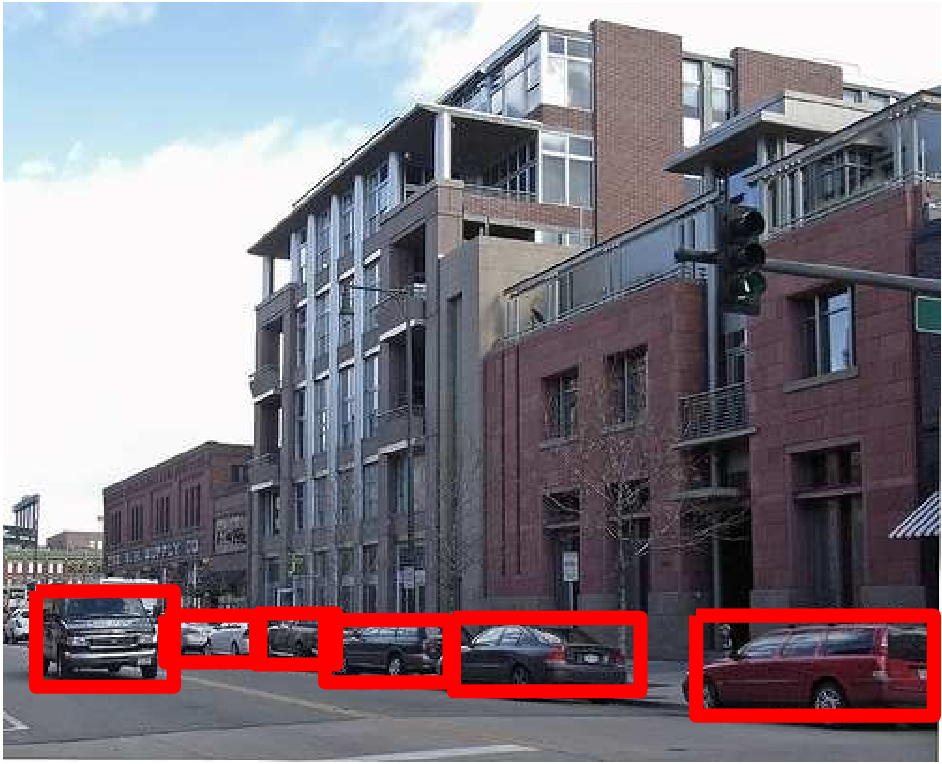} & \
\includegraphics[width=0.32\textwidth]{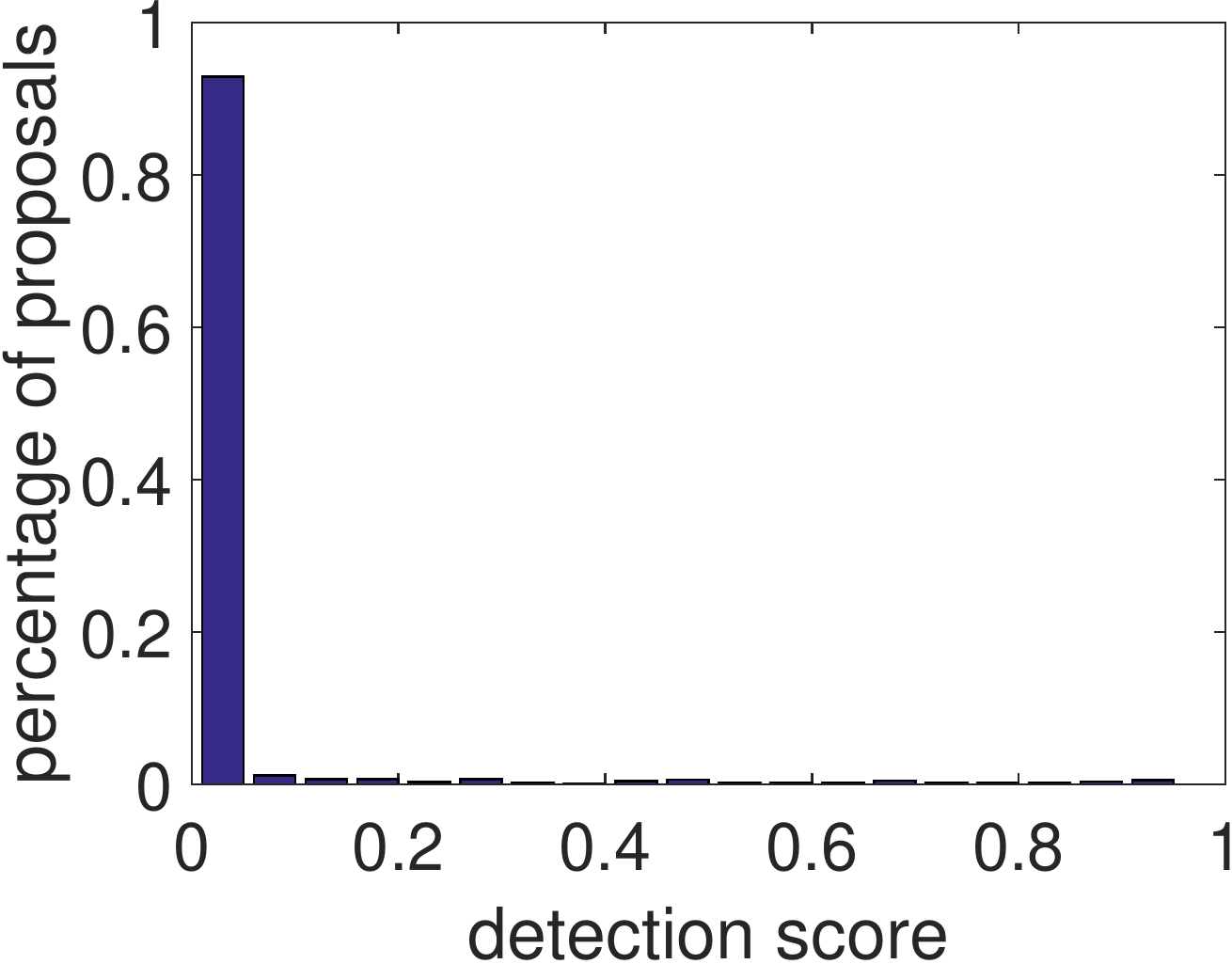} \ \\
(a) & (b)\\
\end{tabular}
\caption{(a) Predicted objects by Fast R-CNN~\cite{DBLP:conf/iccv/Girshick15}. (b) Normalized detection confidence score histogram of object proposals in (a). We observe the same statistics for most images.}
\label{fig:long-tail} %
\end{SCfigure*}

\subsection{The behaviour of an accurate strongly supervised detector}
\label{subsec:long-tail}
Object proposals~\cite{DBLP:journals/ijcv/UijlingsSGS13,DBLP:conf/eccv/ZitnickD14}, which are
image regions that are likely to contain objects, have been widely used in state-of-the-art object detection approaches~\cite{DBLP:journals/pami/GirshickDDM15,DBLP:journals/pami/HeZR015,DBLP:conf/iccv/Girshick15}.
In this section we are interested in the statistics of proposal detection confidence scores on an image generated by a strongly supervised detector.
The observation here motivates our formulation for learning common object detectors in Sec.~\ref{subsec:heatmap}. 

More specifically, we apply one state-of-the-art strongly supervised object detector Fast R-CNN~\cite{DBLP:conf/iccv/Girshick15} (trained on PASCAL VOC 2007 trainval set~\cite{DBLP:journals/ijcv/EveringhamEGWWZ15}) to a PASCAL VOC 2007 test image which contains the object of interest (Fig.~\ref{fig:long-tail} (a)).
After obtaining the detection confidence scores of the more than $2000$ object proposals~\cite{DBLP:journals/ijcv/UijlingsSGS13} extracted from this image, we calculate the normalized histogram of detection 
confidence scores of all proposals (Fig.~\ref{fig:long-tail} (b)). 

From Fig.~\ref{fig:long-tail} (b) it is clear that, although there are multiple 
instances of the object of interest (``car'' in this case), 
more than $90\%$ of object proposals have a very low detection confidence score (less the $0.05$), which indicates that a dominantly large portion of proposals are likely to cover image regions that do not cover the object of interest tightly.
This is understandable as object proposal generation is a pre-processing step in object detection systems,
where recall rate much more important than precision (not missing any objects of interest is more important than generating less false positives).

\subsection{Learning detectors by modeling detection score distribution}
\label{subsec:heatmap}
In the setting of image co-localization, although all we know is that there exists a common object category across images, we still aim to learn the common object detector.
This is possible by modeling the distribution of proposals detection confidence scores. 
More specifically, in our method the common object detector will be learned by enforcing its the distribution of detection confidence scores to mimic that of an accurate strongly supervised detector (Sec.~\ref{subsec:long-tail}).

Formally, for each image $I_i \in \mathcal{I}$, we first extract a set of object proposals
$\mathcal{B}_i=\{B_{i,1},B_{i,2},\ldots,B_{i,M_i}\}$ using EdgeBox~\cite{DBLP:conf/eccv/ZitnickD14}, the performance of which has been illustrated in a recent review~\cite{DBLP:journals/pami/HosangBDS16}.
Let $\feature(B_{i,j}) \in \mathbb{R}^K$ denote the feature representation of proposal $B_{i,j} \in \mathcal{B}_i$.
The particular detection confidence scores that we use are formulated as follows
\begin{equation}
\label{eq:score}
s_{i,j} = f(\w^T\feature(B_{i,j})+b),
\end{equation}
where $\w \in \mathbb{R}^K$, $b \in \mathbb{R}^1$ denote weight and bias terms of the detector respectively,
and $f(\cdot)$ is the softplus function which has the form $f(x) = \ln(1+\exp(x))$.

Irrespective of the form of the detector, we can construct the set of detection confidence scores $\mathcal{S}_i=\{s_{i,1},s_{i,2},\ldots,s_{i,M_i}\}$ over all the proposals $\mathcal{B}_i$ of image $I_i$,
and normalized them as $p_{i,j} = \frac{s_{i,j}+\epsilon}{\sum_{j}(s_{i,j}+\epsilon)}$, where the 
parameter $\epsilon$ is a small constant. 
If the detector in Eq. (\ref{eq:score}) is trained with strong supervision, according to the observation in Sec.~\ref{subsec:long-tail}, most of its detection confidence scores in $\mathcal{S}_i$ should have near-zero values which means that the score vector $\mathbf{s}_i = [s_{i,1}, s_{i,2},\cdots, s_{i,M_i}]^T$ and its normalized version $\mathbf{p}_i = [p_{i,1}, p_{i,2},\cdots, p_{i,M_i}]^T$ should be sparse vectors.
Note that when all proposals have zero detection confidence scores, $\mathbf{s}_i$ will be sparse but $\mathbf{p}_i$ will be dense due to the effect of the constant $\epsilon$. Thus, our method will be based on $\mathbf{p}_i$ because enforcing its sparsity will be equivalent to requiring the detector to have few high detection confidence scores and many low (zero) detection confidence scores, in other words, the detection confidence score distribution will mimic that of an accurate strongly supervised detector.\\

\noindent\textbf{Objective function}.
To measure the sparsity of the normalized detection confidence score vector $\mathbf{p}_i$, we utilize the Shannon entropy as a sparsity indicator, that is, 
\begin{equation}
\label{eq:entropy}
\mathcal{L}(\mathbf{p}_i) = - \sum_{j=1}^{M_i} p_{i,j} \log p_{i,j},
\end{equation}
and the objective for learning the common object detector is formulated as follows:
\begin{equation}
\label{eq:cost_function}
\min_{\w,b} \frac{1}{N} \sum_{i=1}^N \mathcal{L}(\mathbf{p}_i)+\lambda||\w||_{2}^2,
\end{equation}
where we use the square of the $L_2$-norm of $\w$ as a regularizer on the weight vector.

So the optimal value of the weight and bias of the detector is given by:
\begin{equation}
\label{eq:cost_function_full}
\w^*,b^*=\argmin_{\w,b} -\frac{1}{N} \sum_{i=1}^N \sum_{j=1}^M p_{i,j} \log p_{i,j}+\lambda||\w||_{2}^2.
\end{equation}
Note that Eq. (\ref{eq:cost_function_full}) does not involve a set of manually-labeled negative images,
but rather describes the desired form of the detection confidence score distribution.
The learning process also implicitly takes advantage of the chicken-and-egg relationship between object localization and detection: precisely localized object instances are critical for training a good object detector, and objects can be localized more precisely by a well-trained detector.\\
\begin{figure}[t]
\begin{center}
\includegraphics[width=0.9\linewidth]{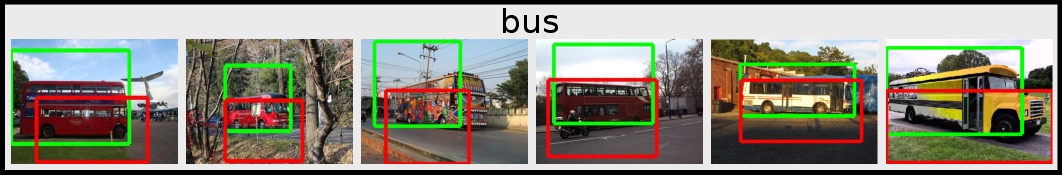} \ \\
\includegraphics[width=0.9\linewidth]{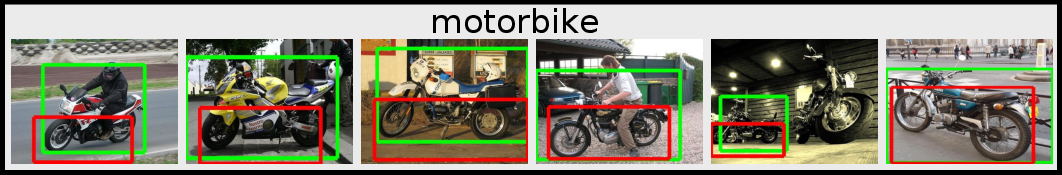} \ \\
\end{center}
\caption{Our detectors fire at different common visual patterns (denoted by red and green bounding boxes) by minimizing Eq.~(\ref{eq:cost_function_full}) with different random initializations.
Although these common visual patterns may not be suitable for the co-localization task, they may be useful for other computer vision tasks, such as 
discovering common object parts for fine-grained image classification~\cite{DBLP:conf/cvpr/KrauseJYL15}.}
\label{fig:multi_fire}
\end{figure}

\noindent\textbf{Optimization}. As our objective function in
Eq. (\ref{eq:cost_function_full}) is non-convex, we minimize it using stochastic gradient descent (SGD).
Similar to the approach used in training a Convolutional Neural Network~\cite{DBLP:conf/nips/KrizhevskySH12}, we divide all data
(\ie, object proposals) into mini-batches.
We initialize the weight vector $\w$ from a zero-mean
Gaussian distribution, while the bias term $b$ is set to zero initially.
During training we divide the learning rate (which is set to $0.1$ initially) by $10$ after each $10$ epochs.
We stop learning after $20$ epochs when the objective function converges. \\

\noindent\textbf{Modification}.
After minimizing Eq. (\ref{eq:cost_function_full}), when we visualize
the proposal with the maximal detection confidence score for each image (Fig.~\ref{fig:multi_fire}), it is interesting to note that the learnt detector may not fire at the common object but some common visual patterns (\eg, common object parts, common object with some context) instead.
Also, the discovered common visual patterns can be very different if the initialization of our detector varies (different local minimums).
However in this work, as we aim to co-localize the common object, we
reformulate Eq. (\ref{eq:score}) by incorporating the ``objectness'' score $o_{i,j}$
(outputs of Edgebox) of each proposal $B_{i,j}$ as a weight to favour proposals with high objectness score (which more likely
to cover a whole object tightly)
\begin{equation}
\label{eq:score_right}
s_{i,j} = o_{i,j}f(\w^T\feature(B_{i,j})+b).
\end{equation}
We experimentally find that minimizing Eq.~(\ref{eq:cost_function_full})  using $s_{i,j}$ defined in Eq.~(\ref{eq:score_right}) gives a stable solution regardless of initialization.\\

\noindent\textbf{Localizing the common object}. The optimal $\w$ and $b$, inserted into Eq.~(\ref{eq:score}), lead to a mechanism for determining the detection confidence scores for all object proposals.  The nature of the co-localization problem means that the maximal score for each image indicates the desired detection.  This method is used as a baseline in the Experiments section (Sec.~\ref{subsec:ablation}).\\

\noindent\textbf{Discussion.} Theoretically, other sparsity measures could be employed to replace the Shannon entropy. Note that the commonly used $L_1$ norm cannot be applied here because $\|\mathbf{p}_i\|_1 = 1$. One possible way to use $L_1$ norm is to redefine the normalization score $p_{i,j} = 
\frac{s_{i,j}+\epsilon}{\sqrt{\sum_{j}(s_{i,j}+\epsilon)^2}}$.

\subsection{Refining the bounding box estimate}
\label{subsec:segmentation}

The quality of the detections generated through the above described process depends entirely on the quality of object proposals.  To overcome this dependency, and enable better final bounding box estimates to be achieved, we have developed a bounding box refinement process as follows.

\begin{figure}[t]
\begin{center}
\begin{tabular}{@{}c@{}c@{}c@{}c@{}c@{}c}
\includegraphics[width=0.15\linewidth, height=0.12\linewidth]{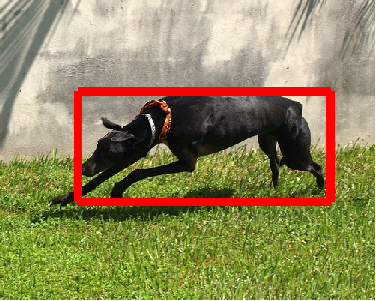} \ &
\includegraphics[width=0.15\linewidth, height=0.12\linewidth]{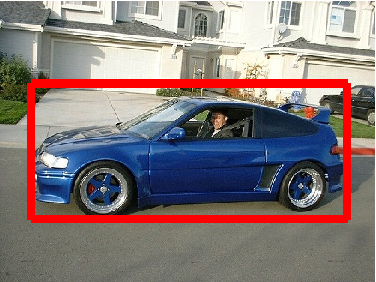} \ &
\includegraphics[width=0.15\linewidth, height=0.12\linewidth]{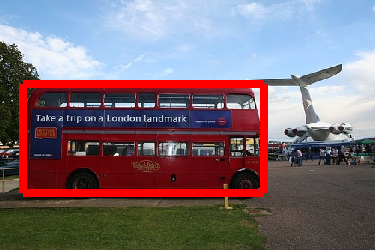} \ &
\includegraphics[width=0.15\linewidth, height=0.12\linewidth]{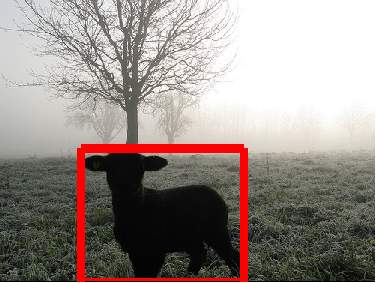} \ &
\includegraphics[width=0.15\linewidth, height=0.12\linewidth]{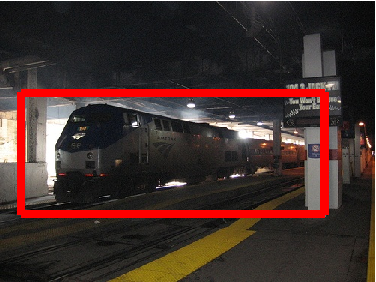} \ &
\includegraphics[width=0.15\linewidth, height=0.12\linewidth]{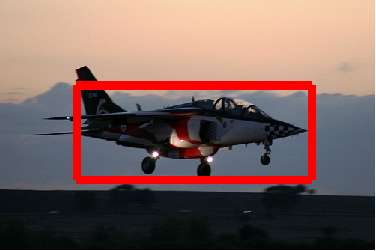} \ \\
\includegraphics[width=0.15\linewidth, height=0.12\linewidth]{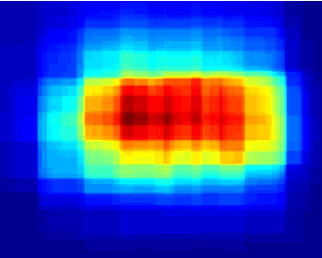} \ &
\includegraphics[width=0.15\linewidth, height=0.12\linewidth]{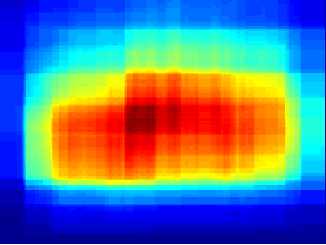} \ &
\includegraphics[width=0.15\linewidth, height=0.12\linewidth]{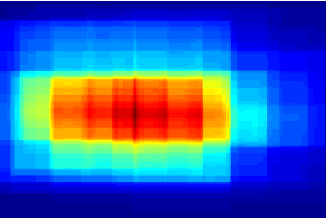} \ &
\includegraphics[width=0.15\linewidth, height=0.12\linewidth]{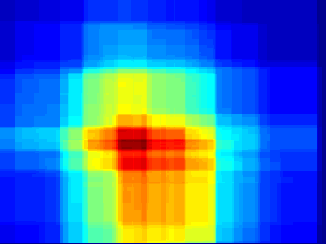} \ &
\includegraphics[width=0.15\linewidth, height=0.12\linewidth]{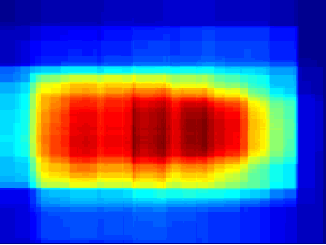} \ &
\includegraphics[width=0.15\linewidth, height=0.12\linewidth]{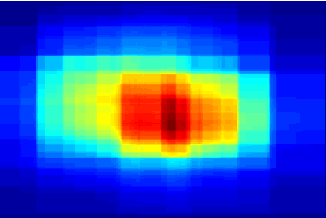} \ \\
\includegraphics[width=0.15\linewidth, height=0.12\linewidth]{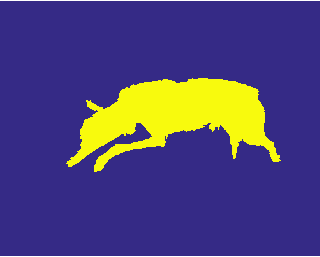} \ &
\includegraphics[width=0.15\linewidth, height=0.12\linewidth]{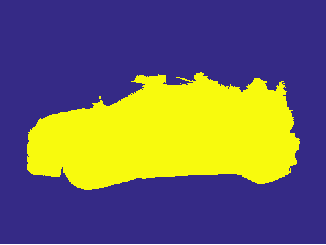} \ &
\includegraphics[width=0.15\linewidth, height=0.12\linewidth]{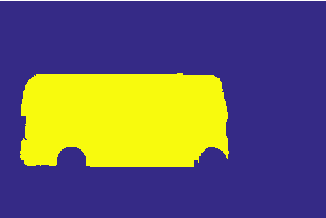} \ &
\includegraphics[width=0.15\linewidth, height=0.12\linewidth]{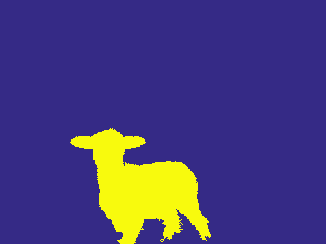} \ &
\includegraphics[width=0.15\linewidth, height=0.12\linewidth]{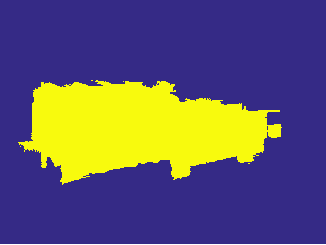} \ &
\includegraphics[width=0.15\linewidth, height=0.12\linewidth]{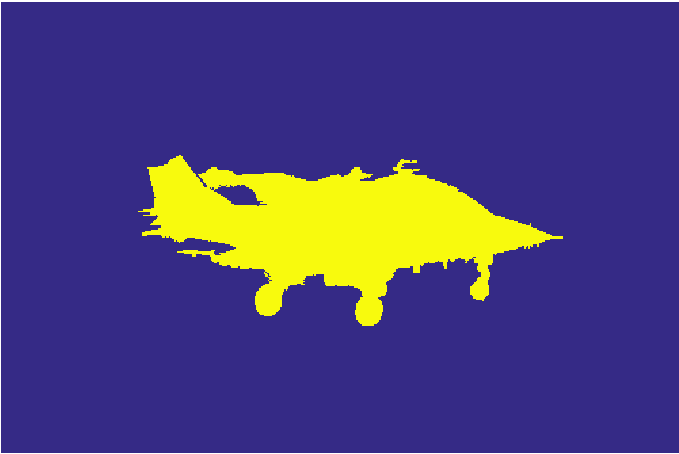} \ \\
\end{tabular}
\end{center}
\caption{Examples of our co-localization process. From top to bottom: input images (predicted boxes in red), detection heat maps, segmentation results. }
\label{fig:seg} %
\end{figure}

Given the optimal $\w^*$ and $b^*$ identified by minimizing
Eq.~(\ref{eq:cost_function_full}), we generate the detection heat map as follows.
For each pixel in the image, we add up the weighted detection confidence score
$s_{i,j}$ from Eq.~(\ref{eq:score_right})
for all proposals $B_{i,j}$ that cover this pixel (zero for pixels not covered by any proposals).
The values are then normalized to the interval $[0,1]$.
This gives rise to a set of detection heat maps $\mathcal{H}=\{H_1,H_2,\ldots,H_n\}$.
Some examples are illustrated in Fig.~\ref{fig:seg}.

Given the set of detection heat maps $\mathcal{H}$, we aim to produce a segmentation of the entire object. This approach is inspired by previous work which casts localization as a segmentation problem (\eg,~\cite{DBLP:conf/iccv/ParkhiVJZ11}).

Formally, we formulate the segmentation problem as a standard graph-cut problem.
We first extract superpixels~\cite{DBLP:journals/ijcv/FelzenszwalbH04} to construct the vertex set $\{m\}$ and  aim to label each superpixel as foreground ($y_m=1$) or background ($y_m=0$).
Mathematically, the energy function is given by
\begin{equation}
\label{eq:energy}
E(\mathbf{y}) = \sum_{m}u_{m}(y_m) + \sum_{(m,n)\in \mathcal{E}}v_{mn}(y_m,y_n),
\end{equation}
where $u_m$ and $v_{mn}$ are the unary and pairwise potential respectively.
$\mathcal{E}$ is the set of edges connecting superpixels\footnote{In our case two superpixels are connected if the distance between their centroids is smaller than the sum of their major axis length.}.\\

\noindent\textbf{Unary potential $u_m$}.
Inspired by~\cite{DBLP:conf/cvpr/KuttelF12}, the unary potential is the novel part of our segmentation framework, which
carries information from the detection heat map $H$:
\begin{equation}
\label{eq:unary}
u_p(y_m) = - \log A_m(y_m),
\end{equation}
where $A_m$ is the prior information from the detection heat map $H$:
\begin{equation}
  \begin{split}
  A_p(y_m=1) & = H(m),  \\
  A_p(y_m=0) & = 1-H(m),
  \end{split}
\end{equation}
where $H(m)$ is the mean of values inside superpixel $m$ on map $H$.\\

\noindent\textbf{Pairwise potential $v_{mn}$}. Our pairwise potential
is defined as follows.
\begin{equation}
\label{eq:pairwise}
v_{mn}(y_m,y_n) = [y_m \neq y_n] e^{-\beta ||C(m)-C(n)||_{2}^2},
\end{equation}
where $C(m)$ is the color histogram feature. 
As in~\cite{DBLP:journals/tog/RotherKB04,DBLP:conf/cvpr/KuttelF12}, this potential penalizes superpixels with different colors taking the same label.\\

As our pairwise potential in Eq.~(\ref{eq:pairwise}) is submodular, the optimal label $\mathbf{y}^*$ can be found efficiently by the graph-cuts~\cite{DBLP:journals/pami/BoykovVZ01}.
As shown in Fig.~\ref{fig:seg}, the segmentation derived through this approach is accurate.
The final bounding box estimate is then calculated as the smallest rectangle which covers the segmentation.

\section{Experiments}
\label{sec:experiments}
\noindent\textbf{Datasets}. We evaluate our approach on three datasets, including PASCAL VOC 2007 and 2012~\cite{DBLP:journals/ijcv/EveringhamEGWWZ15} datasets, six subsets of the ImageNet dataset~\cite{DBLP:conf/cvpr/DengDSLL009} which have not been used in the ILSVRC~\cite{DBLP:journals/ijcv/RussakovskyDSKS15}\footnote{The six categories are chipmunk, rhino, stoat, racoon, rake and wheelchair. Bounding box annotations are available for these categories.}.
For PASCAL VOC datasets, following previous works in co-localization and weakly supervised object localization~\cite{DBLP:conf/cvpr/CinbisVS14,DBLP:conf/cvpr/ChoKSP15,DBLP:conf/cvpr/BilenPT15,DBLP:conf/iccv/WangZYB15}, we use all images
on the \emph{trainval} set discarding images that only contain object instances marked as ``difficult'' or ``truncate''. For ImageNet subsets, 
we filter images with very large ground-truth bounding boxes.\\

\noindent\textbf{Evaluation metric.}
We use two metrics to evaluate our approach.
Firstly, for comparison with state-of-the-art approaches, we use the CorLoc metric~\cite{DBLP:journals/ijcv/DeselaersAF12}, which is defined as the percentage of images that are correctly localized.
An image is considered as correctly localized if the IoU score between the predicted bounding box and any ground-truth bounding boxes of the object of interest exceeds $50\%$.
Instead of using a fixed threshold ($50\%$) for CorLoc metric,
we also compute percentages of correctly localized images under a wide range of thresholds from $0$ to $1$ which results in a \emph{CorLoc curve}.\\

\noindent\textbf{Implementation details}.
We use Edgebox~\cite{DBLP:conf/eccv/ZitnickD14} to extract object proposals with a maximum of $2000$ proposals  extracted from each image.
We represent each Edgebox proposal as a $4096$-dimensional CNN feature from the $fc6$ layer (after ReLU) from the \emph{BVLC Reference CaffeNet} model~\cite{jia2014caffe}. 
We use a fixed value of $1$ for $\lambda$ in Eq.~(\ref{eq:cost_function_full}) which controls the tradeoff between the loss function and regularizer.
The value of $\beta$ in Eq.(\ref{eq:pairwise}) is set to $10$.

\begin{figure*}[t]
\begin{center}
\begin{tabular}{@{}c}
\includegraphics[width=0.9\linewidth]{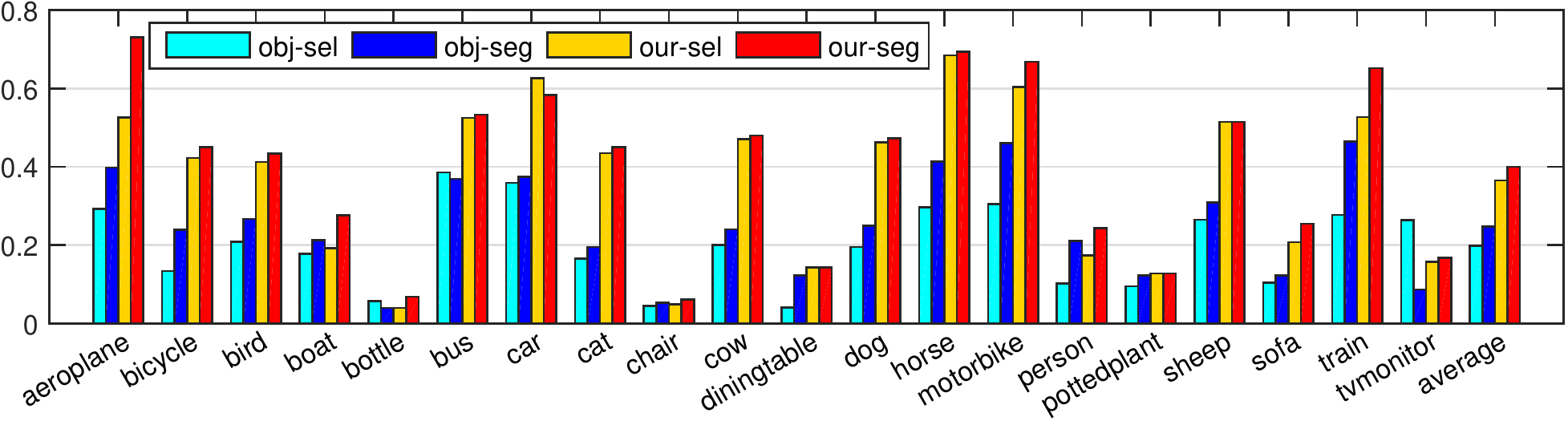} \\
\end{tabular}
\end{center}
\caption{CorLoc scores of our approaches, and baselines, on the PASCAL VOC 2007 dataset.}
\label{fig:baseline}
\end{figure*}

\subsection{Ablation study}
\label{subsec:ablation}
\noindent\textbf{Baselines}. To investigate the impact of the various elements of the proposed approach, we consider the following two baseline methods:
\begin{itemize}
\item ``obj-sel'': the predicted bounding box for an image is simply the proposal with maximum objectness score.
\item ``obj-seg'': for each image, objectness scores of all proposals are treated as detection confidence scores to generate a fake detection heat map, which is then sent to our segmentation model in Sec.~\ref{subsec:segmentation}.
\end{itemize}
The two methods proposed in our work are:
\begin{itemize}
\item ``our-sel'': given the learnt detector~\ref{subsec:heatmap}, we simply select the object proposal which has the maximum detection confidence score $p_{i,j}$ \ie, $B_{i}^* = \argmax_{B_{i,j} \in \mathcal{B}_i}s_{i,j}$. 
\item ``our-seg'': combination of detector training~\ref{subsec:heatmap} and the segmentation refinement~\ref{subsec:segmentation}.
\end{itemize}
Corloc scores for the above four methods on the PASCAL VOC 2007 dataset are illustrated in Fig.~\ref{fig:baseline}.

As shown in Fig.~\ref{fig:baseline}, the simplest baseline ``obj-sel'' does not work well ($19.8\%$ CorLoc). This is because the objectness measure of Edgebox~\cite{DBLP:conf/eccv/ZitnickD14} is heuristically defined based on only edge information.
Therefore, taking the proposal with maximum objectness score is certainly not the optimal method.

However, the ``obj-seg'' baseline in which we use objectness scores to generate a detection heat map for each image, performs quite well, with CorLoc increasing to $24.7\%$.
Surprisingly, this performance is on the par with one state-of-the-art image co-localization approach~\cite{DBLP:conf/eccv/JoulinTF14} ($24.7\%$ \vs  $24.6\%$), even though there is no common object assumption.
This phenomenon indicates that our segmentation model is quite effective.

Thanks to the proposed common object detector learning procedure in Sec.~\ref{subsec:heatmap}, ``our-sel'' achieves a performance of $36.5\%$, outperforming ``obj-sel'' and ``obj-seg'' by over $16\%$ and $11\%$ respectively.
This verifies the effectiveness of this procedure, and particularly that, although we do not have annotated image labels nor bounding boxes, the detector still captures the appearance of the common object, which improves co-localization significantly.

Combing the advantages of the common object detector learning procedure (Sec.~\ref{subsec:heatmap}) and segmentation refinement (Sec.~\ref{subsec:segmentation}), we observe another $3.5\%$ boost in the case of ``our-seg'', reaching $40.0\%$ Corloc. Thus we use ``our-seg'' to compare with state-of-the-art approaches.\\

\noindent\textbf{Number of candidate proposals}.
To test the robustness of our method under different number of candidate object proposals, we select three settings---$500$, $1000$ and $2000$, which result in $39.2\%$, $39.6\%$ and $40.0\%$ CorLoc respectively.
This indicates that our method is quite insensitive to the changes in number of candidate proposals.%

\begin{SCfigure*}[][h]
\centering
\includegraphics[width=0.7\linewidth]{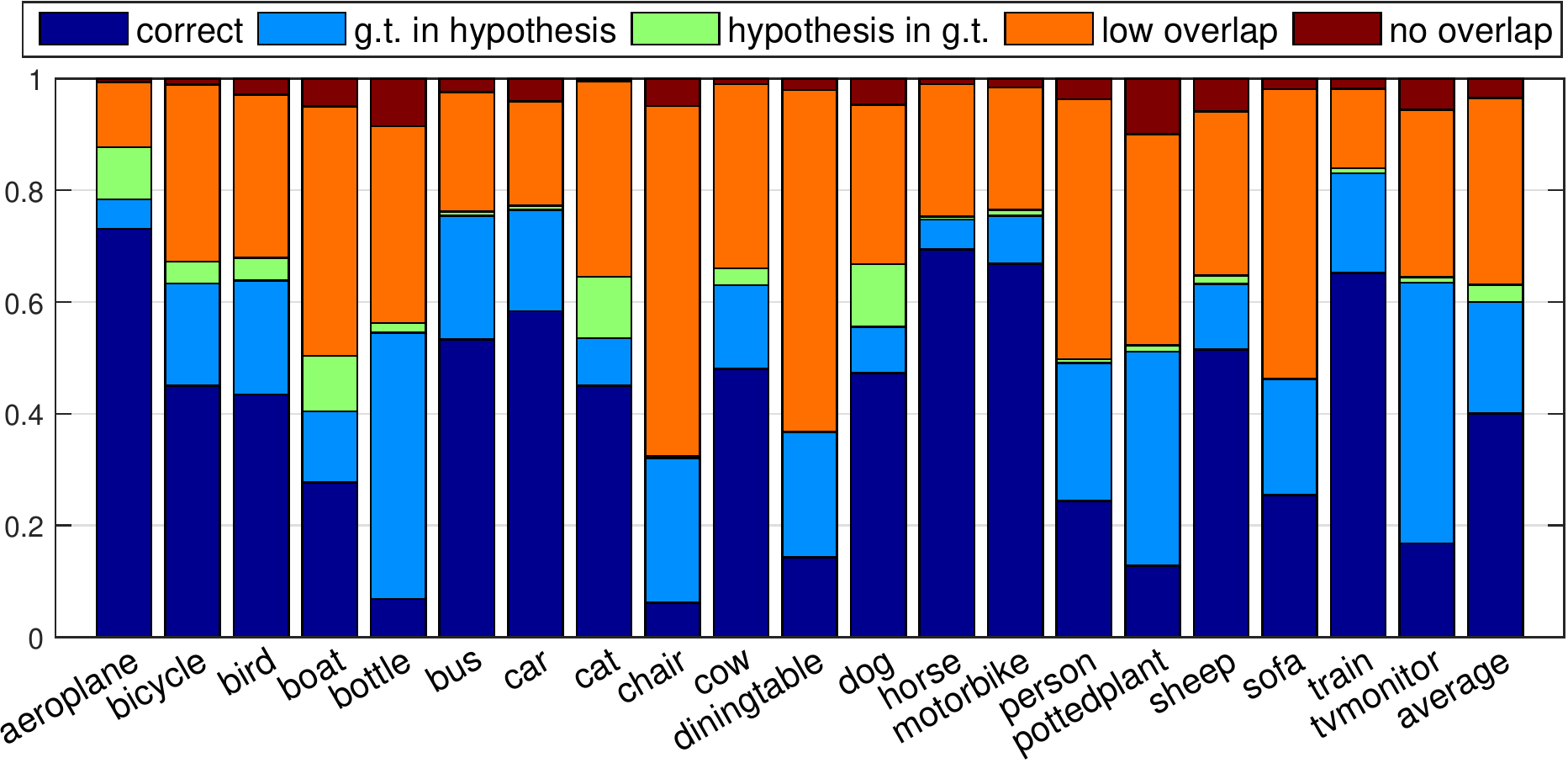} 
\caption{An illustration of error types for our method on the PASCAL VOC 2007 dataset.}
\label{fig:error_mode}
\end{SCfigure*}

\subsection{Diagnosing the localization error}
\label{subsec:diagnosing}
In order to better understand the localization errors, following~\cite{DBLP:conf/eccv/HoiemCD12,DBLP:conf/cvpr/CinbisVS14}, each predicted bounding box predicted by the proposed approach is categorized into the following five cases: (1) correct: IoU score exceeds $50\%$,  (2) g.t. in hypothesis: ground-truth completely inside prediction, (3) hypothesis in g.t.: prediction completely inside ground-truth, (4) no overlap: IoU score equals zero, (5) low overlap:
none of the above four cases.
In Fig.~\ref{fig:error_mode} we show the error modes of our approach across all categories on the PASCAL VOC 2007 dataset.

 As shown in Fig.~\ref{fig:error_mode}, the fraction of ``no overlap'' cases is quite small ($3.5\%$) across all categories, which means our approach can localize common objects to some extent in most cases.
Comparing ``g.t. in hypothesis'' to its ``hypothesis in g.t.'',
it is clear that the former appears more frequently ($19.9\%$ \emph{v.s.} $3.1\%$), which means
our approach  tends to localize objects with some context details.
In terms of correct localization, the three categories with lowest CorLoc values are \emph{bottle} ($6.8\%$), \emph{chair} ($6.2\%$) and \emph{pottedplant} ($12.8\%$).
Images of objects in these categories are always in very clustered environments with
occlusion (\eg, chair is often occluded by table) and large appearance changes.

\subsection{Comparison to state-of-the-art approaches}
\label{subsec:comparison}
\noindent\textbf{Comparison to image co-localization approaches.}
We now compare the results of the proposed approach to the state-of-the-art
image co-localization approaches of Joulin~\etal~\cite{DBLP:conf/eccv/JoulinTF14} and Cho~\etal~\cite{DBLP:conf/cvpr/ChoKSP15} on the PASCAL VOC 2007 dataset
(Table~\ref{tab:voc_2007_colocalization}).
The performance of the proposed method exceeds that of Joulin~\etal~\cite{DBLP:conf/eccv/JoulinTF14}
significantly in most categories, with an improvement of over $15\%$ in mean CorLoc.
The recent method of Cho~\etal~\cite{DBLP:conf/cvpr/ChoKSP15} relies on
matching object parts by Hongh Transform with the predicted bounding box
is selected by a heuristic standout score.
Candidate regions are object proposals represented by whitened HOG features.
However, we found that this whitening process, whose mean vector and covariance matrix are estimated
from the random sampled images from the same dataset (inevitably using images from other categories),
is crucial for the performance of their algorithm.
Our performance bypasses that of \cite{DBLP:conf/cvpr/ChoKSP15} by a reasonable margin of $3.4\%$.

To further verify the effectiveness of the proposed approach, we now present an evaluation on the PASCAL VOC 2012 dataset~\cite{DBLP:journals/ijcv/EveringhamEGWWZ15} which has twice the number of images of PASCAL VOC 2007.
Table~\ref{tab:voc_2012_colocalization} shows our performance along with that of Cho~\etal~\cite{DBLP:conf/cvpr/ChoKSP15} which we evaluated using  their publicly available code.
It is clear that on average our method outperforms that of Cho~\etal~\cite{DBLP:conf/cvpr/ChoKSP15} by $2\%$. \\

\begin{figure*}[t]
\begin{center}
\begin{tabular}{@{}c@{}c@{}c}
\includegraphics[width=0.33\linewidth]{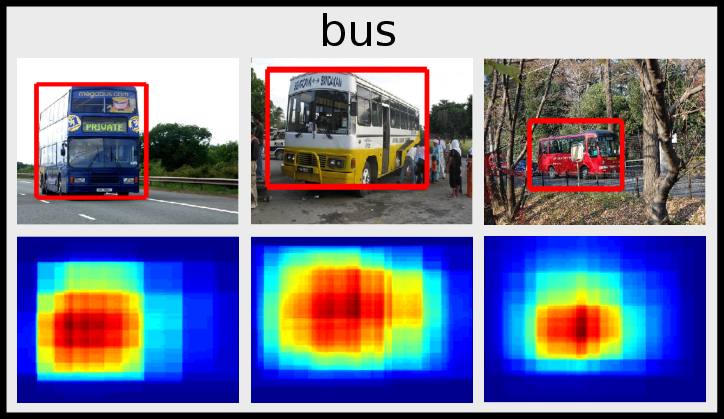} \ &
\includegraphics[width=0.33\linewidth]{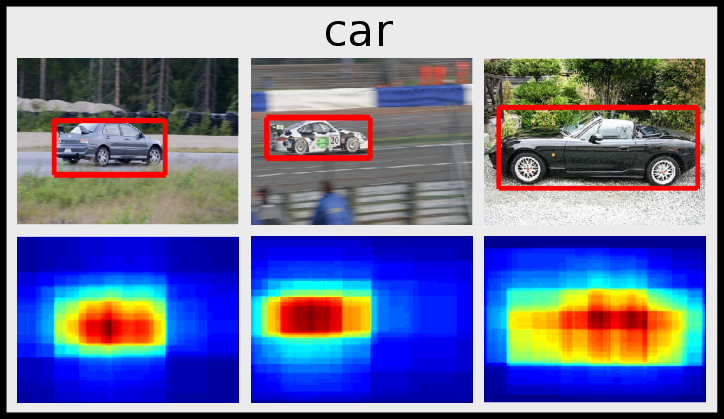} \ &
\includegraphics[width=0.33\linewidth]{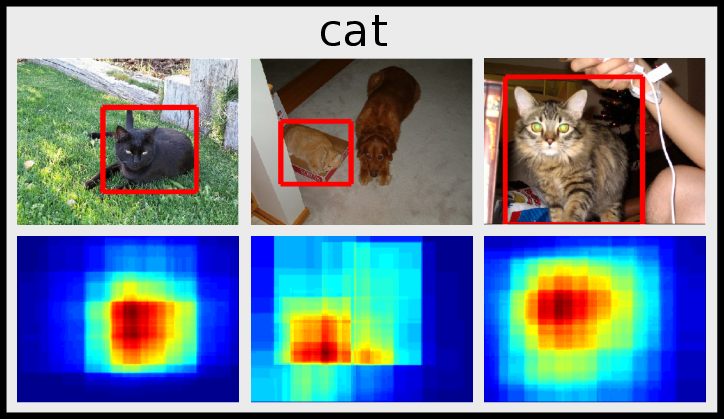} \ \\
\includegraphics[width=0.33\linewidth]{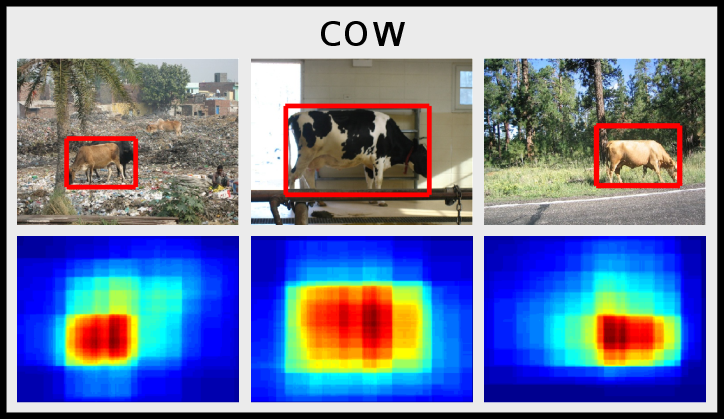} \ &
\includegraphics[width=0.33\linewidth]{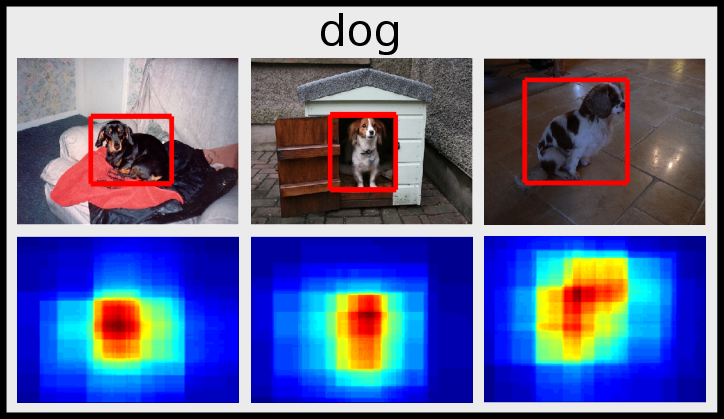} \ &
\includegraphics[width=0.33\linewidth]{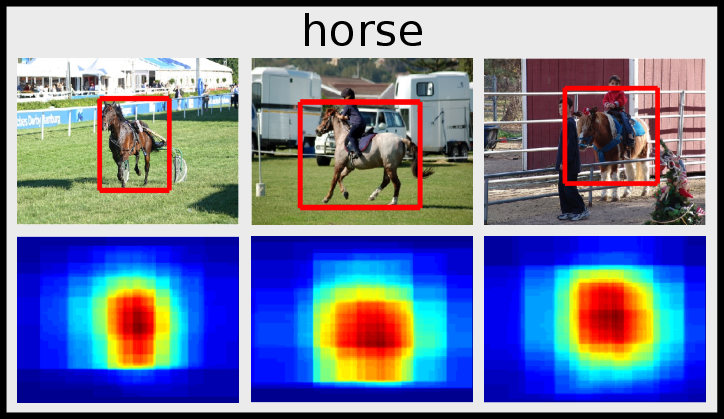} \ \\
\includegraphics[width=0.33\linewidth]{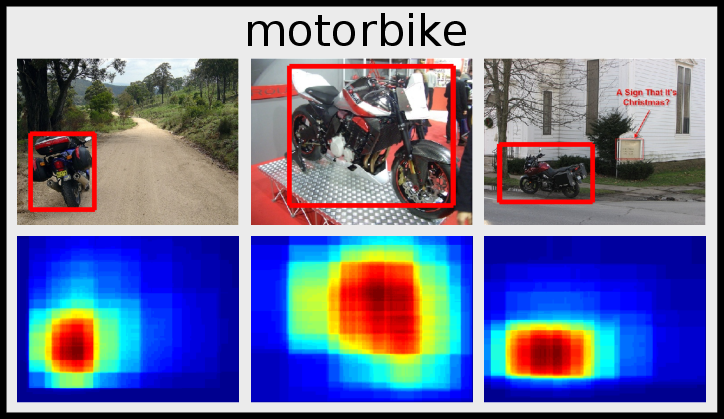} \ &
\includegraphics[width=0.33\linewidth]{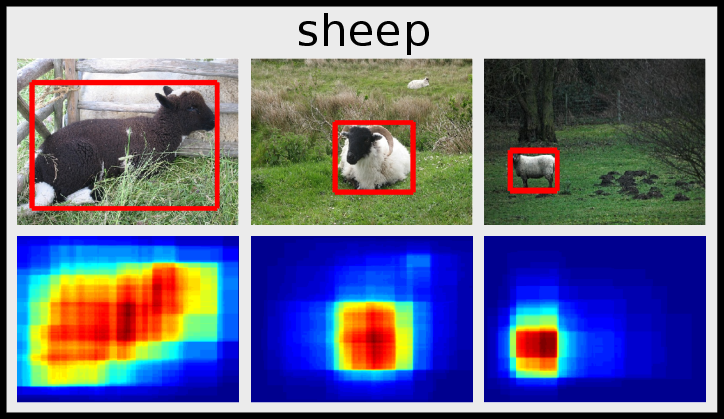} \ &
\includegraphics[width=0.33\linewidth]{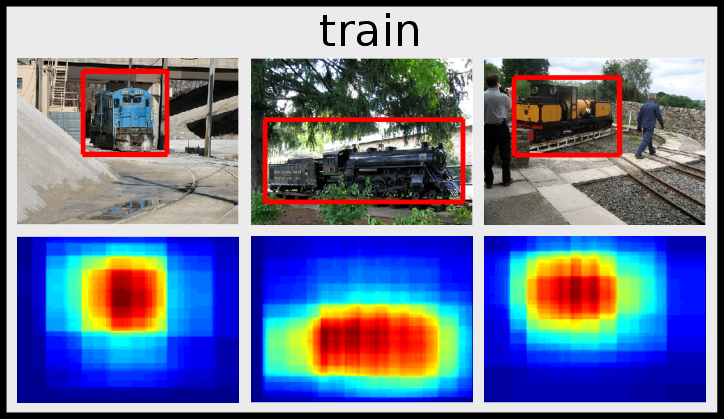} \ \\
\end{tabular}
\end{center}
\caption{Examples of successful co-localization results for the PSACAL VOC 2007 dataset. For each category, the top row depicts predicted bounding boxes on the original image, the bottom row shows corresponding detection heat maps.}
\label{fig:fire_VOC2007}
\end{figure*}

\begin{table*}[t]
\scriptsize{
\setlength{\tabcolsep}{3pt}
\def\arraystretch{1.2}
\center
\caption{Comparison to image co-localization approaches on the PASCAL VOC 2007 dataset in terms of CorLoc metric~\cite{DBLP:journals/ijcv/DeselaersAF12}. }
\scalebox{0.75}{
\begin{tabular}{l@{\hspace{0.4em}}c@{\hspace{0.4em}}c@{\hspace{0.4em}}c@{\hspace{0.4em}}c
@{\hspace{0.4em}}c@{\hspace{0.4em}}c@{\hspace{0.4em}}c@{\hspace{0.4em}}c
@{\hspace{0.4em}}c@{\hspace{0.4em}}c@{\hspace{0.4em}}c@{\hspace{0.4em}}c
@{\hspace{0.4em}}c@{\hspace{0.4em}}c@{\hspace{0.4em}}c@{\hspace{0.4em}}c
@{\hspace{0.4em}}c@{\hspace{0.4em}}c@{\hspace{0.4em}}c@{\hspace{0.4em}}c | c}
\hline
 VOC & aero & bike & bird & boat & bottle & bus & car & cat & chair & cow & table & dog & horse & mbike & person  & plant & sheep & sofa & train & tv & mean\\
\hline
~\cite{DBLP:conf/eccv/JoulinTF14} & 32.8 & 17.3 & 20.9 & 18.2 & 4.5 & 26.9 & 32.7 & 41.0 & 5.8 & 29.1 & \textbf{34.5} & 31.6 & 26.1 & 40.4 & 17.9 & 11.8 & 25.0 & 27.5 & 35.6 & 12.1 & 24.6\\
~\cite{DBLP:conf/cvpr/ChoKSP15} & $50.3$ & $42.8$ & $30.0$ & $18.5$ & $4.0$ & \boldsymbol{$62.3$} & \boldsymbol{$64.5$} & $42.5$ & \boldsymbol{$8.6$} & \boldsymbol{$49.0$} & $12.2$ &
$44.0$ & $64.1$ & $57.2$ & $15.3$ & $9.4$ & $30.9$ & \boldsymbol{$34.0$} & $61.6$ & \boldsymbol{$31.5$} & $36.6$\\
Ours & \boldsymbol{$73.1$} & \boldsymbol{$45.0$} & \boldsymbol{$43.4$} & \boldsymbol{$27.7$} & \boldsymbol{$6.8$} & $53.3$ & $58.3$ & \boldsymbol{$45.0$} & $6.2$ & $48.0$ & $14.3$ & \boldsymbol{$47.3$} & \boldsymbol{$69.4$} & \boldsymbol{$66.8$} & \boldsymbol{$24.3$} & \boldsymbol{$12.8$} & \boldsymbol{$51.5$} & $25.5$ & \boldsymbol{$65.2$} & $16.8$ & \boldsymbol{$40.0$}\\
\hline
\end{tabular}
}
\label{tab:voc_2007_colocalization}
}
\end{table*}

\begin{table*}[t]
\scriptsize{
\setlength{\tabcolsep}{3pt}
\def\arraystretch{1.2}
\center
\caption{Comparison to image co-localization approaches on the PASCAL VOC 2012 dataset in terms of CorLoc metric~\cite{DBLP:journals/ijcv/DeselaersAF12}. }
\scalebox{0.75}{
\begin{tabular}{l@{\hspace{0.4em}}c@{\hspace{0.4em}}c@{\hspace{0.4em}}c@{\hspace{0.4em}}c
@{\hspace{0.4em}}c@{\hspace{0.4em}}c@{\hspace{0.4em}}c@{\hspace{0.4em}}c
@{\hspace{0.4em}}c@{\hspace{0.4em}}c@{\hspace{0.4em}}c@{\hspace{0.4em}}c
@{\hspace{0.4em}}c@{\hspace{0.4em}}c@{\hspace{0.4em}}c@{\hspace{0.4em}}c
@{\hspace{0.4em}}c@{\hspace{0.4em}}c@{\hspace{0.4em}}c@{\hspace{0.4em}}c | c}
\hline
\textbf{VOC}
 & aero & bike & bird & boat & bottle  &  bus  &  car  &  cat  &  chair & cow & table &  dog  & horse & mbike & person  & plant & sheep & sofa & train & tv & mean\\
\hline
\cite{DBLP:conf/cvpr/ChoKSP15} & $57.0$ & $41.2$ & $36.0$ & $26.9$ & $5.0$ & \boldsymbol{$81.1$} & \boldsymbol{$54.6$} & \boldsymbol{$50.9$} & \boldsymbol{$18.2$} & $54.0$ & \boldsymbol{$31.2$} & $44.9$ & $61.8$ & $48.0$ & $13.0$ & $11.7$ & $51.4$ & \boldsymbol{$45.3$} & $64.6$ & \boldsymbol{$39.2$} & $41.8$\\
Ours & \boldsymbol{$65.7$} & \boldsymbol{$57.8$} & \boldsymbol{$47.9$} & \boldsymbol{$28.9$} & \boldsymbol{$6.0$} & $74.9$ & $48.4$ & $48.4$ & $14.6$ & \boldsymbol{$54.4$} & $23.9$ & \boldsymbol{$50.2$} & \boldsymbol{$69.9$} & \boldsymbol{$68.4$} & \boldsymbol{$24.0$} & \boldsymbol{$14.2$} & \boldsymbol{$52.7$} & $30.9$ & \boldsymbol{$72.4$} & $21.6$ & \boldsymbol{$43.8$}\\
\hline
\end{tabular}
}
\label{tab:voc_2012_colocalization}
}
\end{table*}

\begin{table*}[t]
\scriptsize{
\setlength{\tabcolsep}{3pt}
\def\arraystretch{1.2}
\center
\caption{Comparison to weakly supervised object localization approaches on 
the PASCAL VOC 2007 dataset in terms of CorLoc metric~\cite{DBLP:journals/ijcv/DeselaersAF12}. Note that these comparators require access to a negative image set, whereas our method does not.
}
\scalebox{0.75}{
\begin{tabular}{l@{\hspace{0.4em}}c@{\hspace{0.4em}}c@{\hspace{0.4em}}c@{\hspace{0.4em}}c
@{\hspace{0.4em}}c@{\hspace{0.4em}}c@{\hspace{0.4em}}c@{\hspace{0.4em}}c
@{\hspace{0.4em}}c@{\hspace{0.4em}}c@{\hspace{0.4em}}c@{\hspace{0.4em}}c
@{\hspace{0.4em}}c@{\hspace{0.4em}}c@{\hspace{0.4em}}c@{\hspace{0.4em}}c
@{\hspace{0.4em}}c@{\hspace{0.4em}}c@{\hspace{0.4em}}c@{\hspace{0.4em}}c | c}
\hline
\textbf{VOC} & aero  &   bike &  bird & boat &  bottle  &  bus  &  car  &  cat  &  chair & cow & table &  dog  & horse & mbike & person  & plant & sheep & sofa & train & tv & mean\\
\hline
\cite{DBLP:conf/iccv/SivaX11} & $42.4$ & $46.5$ & $18.2$ & $8.8$ & $2.9$ & $40.9$ & $73.2$ & $44.8$ & $5.4$ & $30.5$ & $19.0$ & $34.0$ & $48.8$ & $65.3$ & $8.2$ & $9.4$ & $16.7$ & $32.3$ & $54.8$ & $5.5$ & $30.4$\\
\cite{DBLP:conf/iccv/ShiHX13} & $67.3$ & $54.4$ & $34.3$ & $17.8$
& $1.3$ & $46.6$ & $60.7$ & \boldsymbol{$68.9$} & $2.5$ & $32.4$ & $16.2$ & \boldsymbol{$58.9$} & $51.5$ & $64.6$ & $18.2$ & $3.1$ & $20.9$ & $34.7$ & $63.4$ & $5.9$ & $36.2$\\
\cite{DBLP:conf/cvpr/CinbisVS14} & $56.6$ & $58.3$ & $28.4$ & $20.7$ & $6.8$ & $54.9$ & $69.1$ & $20.8$ & $9.2$ & $50.5$ & $10.2$ & $29.0$ & $58.0$ & $64.9$ & $36.7$ & $18.7$ & $56.5$ & $13.2$ & $54.9$ & $59.4$ & $38.8$\\
\cite{DBLP:conf/iccv/WangZYB15} & $37.7$ & $58.8$ & $39.0$ & $4.7$ & $4.0$ & $48.4$ & $70.0$ & $63.7$ & $9.0$ & $54.2$ & \boldsymbol{$33.3$} & $37.4$
& $61.6$ & $57.6$ & $30.1$ & $31.7$ & $32.4$ & \boldsymbol{$52.8$} & $49.0$ & $27.8$ &
$40.2$\\
\cite{DBLP:conf/cvpr/BilenPT15} & $66.4$ & $59.3$ & $42.7$ & $20.4$ & \boldsymbol{$21.3$} & \boldsymbol{$63.4$} & \boldsymbol{$74.3$} & $59.6$ & $21.1$ & $58.2$ & $14.0$ & $38.5$ & $49.5$ & $60.0$ & $19.8$ & $39.2$ & $41.7$ & $30.1$ & $50.2$ & $44.1$ & $43.7$\\
\cite{DBLP:journals/pami/RenHTT16} & $79.2$ & $56.9$ & $46.0$ & $12.2$ & $15.7$ & $58.4$ & $71.4$ & $48.6$ & $7.2$ & \boldsymbol{$69.9$} & $16.7$ & $47.4$ & $44.2$ & \boldsymbol{$75.5$} & \boldsymbol{$41.2$} & \boldsymbol{$39.6$} & $47.4$ & $32.2$ & $49.8$ & $18.6$ & $43.9$\\
\cite{DBLP:conf/eccv/WangRHT14} & \boldsymbol{$80.1$} & \boldsymbol{$63.9$} & \boldsymbol{$51.5$} & $14.9$ & $21.0$ &
$55.7$ & $74.2$ & $43.5$ & \boldsymbol{$26.2$} & $53.4$ & $16.3$ & $56.7$ & $58.3$ & $69.5$ & $14.1$ & $38.3$ & \boldsymbol{$58.8$} & $47.2$ & $49.1$ & \boldsymbol{$60.9$} & \boldsymbol{$48.5$}\\

Ours & $73.1$ & $45.0$ & $43.4$ & \boldsymbol{$27.7$} & $6.8$ & $53.3$ & $58.3$ & $45.0$ & $6.2$ & $48.0$ & $14.3$ & $47.3$ & \boldsymbol{$69.4$} & $66.8$ & $24.3$ & $12.8$ & $51.5$ & $25.5$ & \boldsymbol{$65.2$} & $16.8$ & $40.0$\\
\hline
\end{tabular}
}
\label{tab:voc_2007_wsl}
}
\end{table*}

\begin{table*}[b]
\scriptsize{
\setlength{\tabcolsep}{3pt}
\def\arraystretch{1.2}
\caption{Comparison to image co-localization approaches on the ImageNet subsets in terms of CorLoc metric~\cite{DBLP:journals/ijcv/DeselaersAF12}. Note that these categories have not been used for pre-training the CNN model, which is used as a feature extractor in this work.}
\begin{center}
\begin{tabular}{l@{\hspace{0.6em}}c@{\hspace{0.6em}}c@{\hspace{0.6em}}c@{\hspace{0.6em}}c
@{\hspace{0.6em}}c@{\hspace{0.6em}}c | c}
\hline
\textbf{ImageNet}
 & chipmunk & rhino & stoat & racoon & rake & wheelchair & mean\\
\hline
Cho \etal~\cite{DBLP:conf/cvpr/ChoKSP15} & $26.6$ & \boldsymbol{$81.8$} & $44.2$ & $30.1$ & $8.3$ & $35.3$ & $37.7$\\
Ours & \boldsymbol{$44.9$} & \boldsymbol{$81.8$} & \boldsymbol{$67.3$} & \boldsymbol{$41.8$} & \boldsymbol{$14.5$} & \boldsymbol{$39.3$} & \boldsymbol{$48.3$}\\
\hline
\end{tabular}
\end{center}
\label{tab:imagenet_colocalization}
}
\end{table*}

\begin{figure*}[t]
\begin{center}
\begin{tabular}{@{}c@{}c@{}c}
\includegraphics[width=0.33\linewidth]{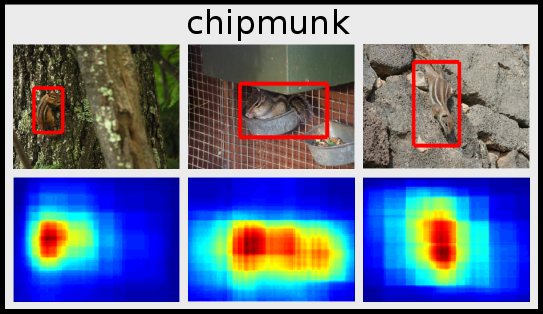} \ &
\includegraphics[width=0.33\linewidth]{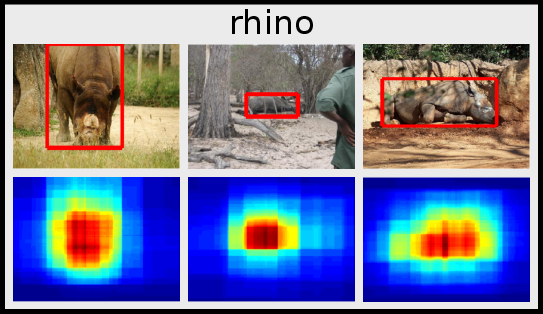} \ &
\includegraphics[width=0.33\linewidth]{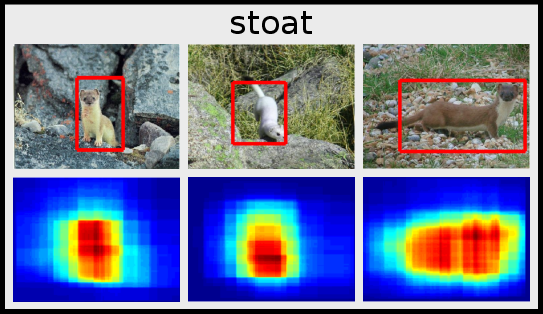} \ \\
\includegraphics[width=0.33\linewidth]{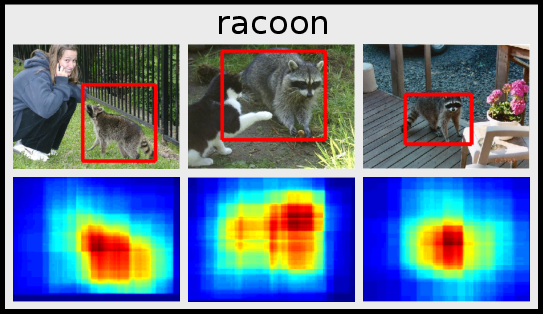} \ &
\includegraphics[width=0.33\linewidth]{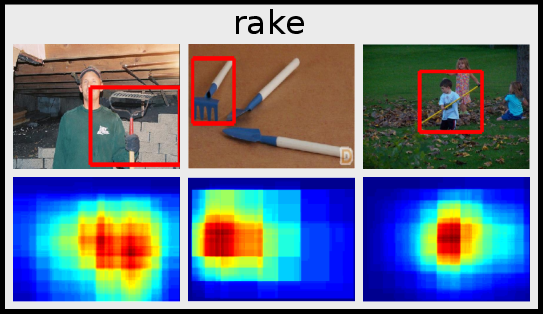} \ &
\includegraphics[width=0.33\linewidth]{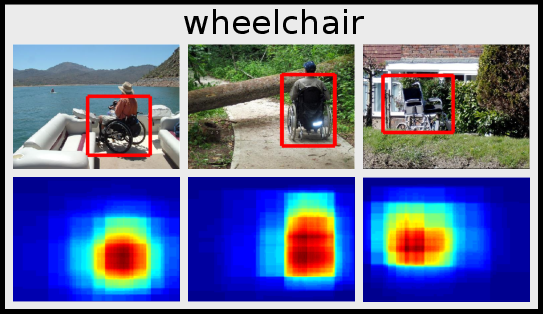} \ \\
\end{tabular}
\end{center}
\caption{Examples of successful co-localization results for the ImageNet subsets. }
\label{fig:fire_imagenet}
\end{figure*}

\begin{figure*}[h]
\begin{center}
\begin{tabular}{@{}c@{}c}
\includegraphics[width=0.5\linewidth]{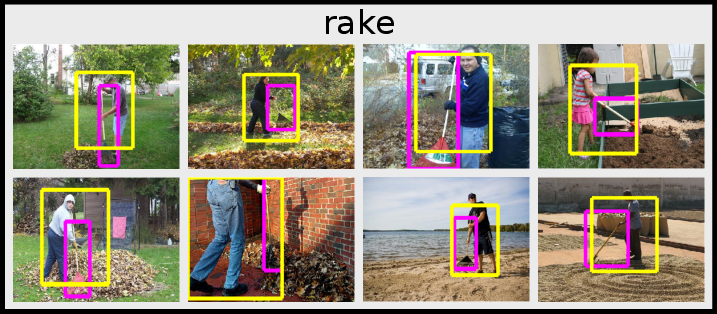} \ &
\includegraphics[width=0.5\linewidth]{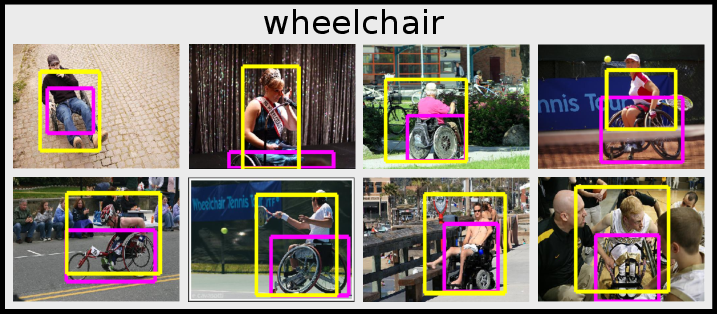} \ \\
\end{tabular}
\end{center}
\caption{Examples of failure cases of \emph{rake} and \emph{wheelchair} (ground truth in megenta and predicted boxes in yellow).}
\label{fig:fire_imagenet_fail}
\end{figure*}

\noindent\textbf{Comparison to weakly supervised object localization approaches.} We also compare the proposed approach with some state-of-the-art approaches
on weakly supervised object localization.
Table~\ref{tab:voc_2007_wsl} illustrates the comparison of several recent works and our approach on PASCAL VOC 2007 dataset.
In particular, our performance ($40.0\%$) is comparable to that of a very recent work~\cite{DBLP:conf/iccv/WangZYB15}($40.2\%$) which also uses CNN features and Edgebox proposals.
Please note that in our framework we do not have any negative images whereas WSL approaches do, which means the we are
addressing a more challenging problem.
As shown in Table~\ref{tab:voc_2007_wsl}, though we do not have any negative images, we still outperforms WSOL approaches on $3$ of $20$ categories.

We have also conducted an object detection experiment on PASCAL VOC 2007. Specifically, for each category, we treated predicted bounding boxes of our co-localization algorithm on trainval set as ground-truth annotations and sampled proposals from other categories or have a overlap ratio less than $0.1$ against our localized bounding boxes as negative samples. The fc6 feature from the CaffeNet are extracted and hard negative mining is performed to train the detector. We achieve a mAP of $16.7\%$ on the testset when using a nms threshold of $0.5$. Although our performance is lower than some WSOL methods, it is understandable as we do not use negative data for co-localization. Moreover, we can easily extend our formulation (Eq.\ref{eq:cost_function_full}) to handle negative data and thus perform WSOL.\\

\noindent\textbf{Visualization.}
In Fig.~\ref{fig:fire_VOC2007}, we provide a set of successful co-localization results along with the corresponding detection heat maps for some categories of the PASCAL VOC 2007 dataset.
It demonstrates that detection heat maps successfully predict the correct location of the common object regardless of changes in scale, appearance and viewpoint. This provides a strong indication that, although trained without annotated positive or negative examples, our method is able to
discriminate the common object from other objects in the scene.
\subsection{ImageNet subsets}
We note that the CNN model used for extracting features is pre-trained in the ILSVRC~\cite{DBLP:journals/ijcv/RussakovskyDSKS15}, whose training set may have some overlapping categories with the VOC datasets.
In order to justify the proposed method is insensitive to the object category, we randomly selected six subsets of the ImageNet~\cite{DBLP:conf/cvpr/DengDSLL009} which have not been used in the ILSVRC (thus ``unseen'' by the CNN model) for evaluation. 

Table~\ref{tab:imagenet_colocalization} shows our co-localization result 
along with that of the current state-of-the-art work of Cho~\etal~\cite{DBLP:conf/cvpr/ChoKSP15}. Clearly, the proposed approach outperforms~\cite{DBLP:conf/cvpr/ChoKSP15} by a reasonable margin on all categories except the \emph{rhino} category, whose images tend to have 
relatively large common instances and less cluttered background.
Some successfully co-localization samples are depicted in Fig.~\ref{fig:fire_imagenet}.

We also visualize some failure cases of the two categories our approach 
performed worst---\emph{rake} and \emph{wheelchair} ( Fig.~\ref{fig:fire_imagenet_fail}).
Interestingly, these failure cases are quite understandable. 
For example, a large portion of images in the \emph{rake} category have the 
scenario in which people are holding a rake, thus our co-localization approach tends to capture this scenario as the ``common object''. 
A similar phenomenon is also observed in the \emph{wheelchair} category in which people tend to sit on the wheelchair.

\section{Conclusion}
\label{conclusion}
We have addressed the image co-localization problem by directly learning a common object detector. The key discovery made in this paper is that this detector can be learned with the objective of making its detection score distribution mimic an accurate strongly supervised object detector. Also, we have illustrated that it is profitable to use a CRF model to refine the co-localization result, which has not been explored in recent works on co-localization.

\noindent{\bf Acknowledgements}
This work was in part supported by the Data to Decisions CRC Centre.
C. Shen's participation was in part supported by ARC Future Fellowship No.\ FT120100969.

\bibliographystyle{splncs03}
\bibliography{egbib}
\end{document}